\documentclass[11pt]{article}

\usepackage[preprint]{acl}

\usepackage{times}
\usepackage{latexsym}

\usepackage[T1]{fontenc}
\usepackage[utf8]{inputenc}

\usepackage{microtype}

\usepackage{inconsolata}

\usepackage{graphicx}
\usepackage{inconsolata}
\usepackage{graphicx} 
\usepackage{booktabs} 
\usepackage{multirow} 
\usepackage{amsmath}
\usepackage{amssymb}
\usepackage{stfloats}
\usepackage{float}
\usepackage{bbm}
\usepackage{algorithm}
\usepackage{algpseudocode}
\usepackage{tikz}
\usepackage{pgfplots}
\usepgfplotslibrary{groupplots}
\pgfplotsset{compat=1.18}

\title{VisER: Visual Evidence and Reliance \\for Object Hallucination Detection in LVLMs}

\author{
Afsaneh Hasanebrahimi,
Hanxun Huang,
Christopher Leckie,
Sarah Erfani \\
School of Computing and Information Systems \\
The University of Melbourne \\
Melbourne, Australia \\
\texttt{\{a.hasanebrahimi,hanxun.huang,caleckie,sarah.erfani\}@unimelb.edu.au}
}

\begin{document}
\maketitle

\begin{abstract}
Object hallucination remains a persistent reliability issue in large vision-language models, where generated object mentions may sound plausible but lack visual grounding. Recent training-free detectors use internal signals such as token likelihood, attention, visual confidence, or image-text similarity to identify hallucinated objects. %These signals are useful, but they mostly give a one-sided view of grounding by measuring how strongly an object is supported inside the model.
These signals are useful, but they are often source-confounded. They measure how strongly an object is supported inside the model without distinguishing whether that support comes from object-specific visual evidence or the generated text prefix. In difficult cases, a hallucinated object can still receive high internal support because it fits the scene, is associated with nearby visual cues, or follows naturally from the generated text prefix. We propose VisER, a training-free two-sided metric for object-level hallucination detection. VisER evaluates each generated object mention from two complementary views. Visual Evidence measures whether object-context compatibility is backed by object-specific evidence from image tokens. Visual Reliance measures whether the object is supported more by the image than by the generated prefix. Combining these views gives a more source-aware grounding score, while avoiding additional object-level verification generations. Across multiple LVLMs and benchmarks, VisER improves AUROC and AUPR over a range of baselines.

\end{abstract}

\section{Introduction}
Large Vision-Language Models (LVLMs) have made substantial progress in connecting visual perception with natural language generation, enabling strong performance across image captioning, visual question answering, and multimodal instruction following~\citep{liu2023visual,zhu2024minigpt,dai2023instructblip,chen2023shikra,zhu2025internvl3}. Despite these advances, LVLMs remain vulnerable to object hallucination, where the model generates object mentions that are not present in the input image~\citep{rohrbach2018object,li2023evaluating}. This failure mode is particularly concerning because hallucinated objects are often fluent, contextually plausible, and embedded naturally within otherwise correct descriptions. As LVLMs are increasingly used in settings where factual visual grounding is important, detecting object hallucinations has become a key problem for reliable multimodal generation~\citep{bai2024hallucination,liu2024survey}.

Existing object hallucination detectors can be broadly grouped into two categories. The first relies on external supervision, including ground-truth object annotations, reference-based evaluation, or auxiliary judge models~\citep{rohrbach2018object,li2023evaluating,jing2024faithscore,xiao2025detecting,sun2024aligning}. While effective for controlled evaluation, these methods are difficult to deploy in open-ended settings where references are unavailable, and auxiliary evaluators may introduce their own biases or errors. The second category uses training-free internal signals available during LVLM inference, such as token probabilities~\citep{zhou2024analyzing}, confidence estimates from visual representations~\citep{jiang2025interpreting}, attention patterns~\citep{jiang2025devils,hoang2026pas}, and latent image-text similarities~\citep{phukan2025beyond,park2025glsim}. These methods are lightweight and reference-free, but they mainly ask whether an object mention is supported inside the model. %In difficult cases, however, support strength alone may be insufficient, because the same signal can increase for different reasons.

\begin{figure*}[t]
    \centering
    \includegraphics[width=\textwidth]{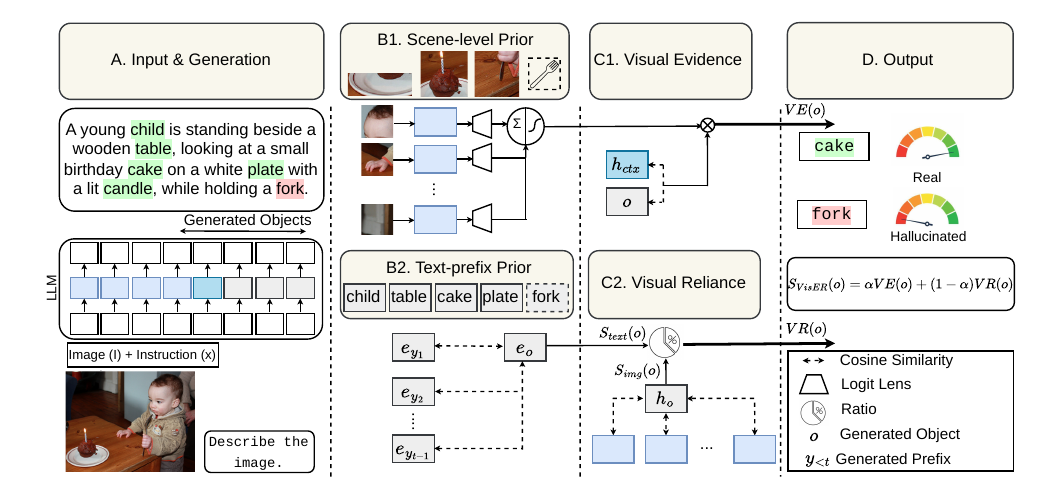}
    \caption{
    Overview of VisER as two-sided visual source validation. A generated object mention can receive high internal support either because it is visually grounded or because it is falsely supported by scene priors, misleading visual cues, or autoregressive text-prefix continuation. VisER validates the source of this support using two complementary signals: Visual Evidence \(VE(o)\), which checks whether object-context compatibility is backed by object-specific image-token evidence, and Visual Reliance \(VR(o)\), which compares image-derived support with generated-prefix support. The final score favors objects that are both visually evidenced and image-driven.
    }
    \label{fig:pipeline}
\end{figure*}

The need for a two-sided view becomes clear in difficult hallucination cases. Under the standard evaluation definition, an object hallucination occurs when a generated object mention is absent from the image~\citep{rohrbach2018object,li2023evaluating}. However, for a detector, the difficult cases are those where the absent object still receives strong internal support. Scene plausibility and object co-occurrence can make an absent object appear compatible with the global visual context or the surrounding generated description~\citep{rohrbach2018object,li2023evaluating,zhou2024analyzing,park2025glsim}. Autoregressive generation can also reinforce later object mentions through the generated prefix rather than through the image~\citep{zhou2024analyzing,huang2024opera,liu2024paying,hoang2026pas}. Even image-side signals can be misleading when attention or similarity is concentrated on related, background, or non-object-specific visual tokens~\citep{gong2024damro,jiang2025devils}. Thus, high likelihood, attention, confidence, or image-text similarity can indicate visual grounding, but it can also indicate plausibility or association. 

We refer to this ambiguity as \emph{source confounding}; an internal signal is source-confounded when it assigns high support to an object mention without distinguishing the source of that support. For a grounded object, high support should come from object-specific visual evidence. For a hallucinated object, however, high support may instead arise because the object is scene-plausible, associated with related visual cues, or naturally continued from the generated text prefix. In such cases, the detector measures the strength of support but not whether that support is visually grounded. This makes source confounding a central failure mode for training-free hallucination detection, and a reliable detector should not only ask whether an object is internally supported, but also whether the support is object-specific and image-derived. Appendix~\ref{app:qualitative} shows representative examples in which contextually plausible but absent objects are scored as grounded by baseline signals, whereas VisER correctly flags them as hallucinated.

Motivated by this observation, we propose \textbf{VisER} (\textbf{Vi}sual \textbf{E}vidence and \textbf{R}eliance), a training-free metric for object-level hallucination detection. As illustrated in Figure~\ref{fig:pipeline}, VisER evaluates each generated object mention from two complementary sides. \textit{Visual Evidence} measures whether object-context compatibility is backed by object-specific evidence from image tokens. \textit{Visual Reliance} measures whether the object is supported more by the image than by the generated text prefix. Combining these two sides yields a more source-aware grounding score without model training or additional object-level verification generations. Unlike similarity-based detectors that mainly measure compatibility between an object and the image-text context~\citep{park2025glsim}, VisER further asks whether that compatibility is supported by object-specific visual evidence and is not primarily driven by the prefix.

% Experiments across multiple LVLMs and benchmarks show that VisER consistently improves training-free object hallucination detection over likelihood-based, attention-based, and similarity-based baselines. Our ablations further show that Visual Evidence Support and Source Reliance address complementary failure modes: the former suppresses prior-driven compatibility that lacks object-specific evidence, while the latter identifies object mentions whose support is better explained by the generated text prefix than by the image.

Our contributions are summarized as follows:
\begin{itemize}
\item We revisit training-free object hallucination detection through source confounding, where high internal support may reflect visual evidence, scene association, or generated-prefix cues.

\item We propose VisER, a training-free metric that combines Visual Evidence and Visual Reliance to score object mentions based on both object-specific evidence and image-versus-prefix support.

\item We evaluate VisER across multiple LVLMs and benchmarks, showing consistent gains over existing baselines and complementary benefits from both components.
\end{itemize}

\section{Related Work}
\paragraph{Object hallucination evaluation and external verification.}
Object hallucination refers to a visual faithfulness error in which an LVLM mentions objects that are absent from, or unsupported by, the input image~\citep{liu2024survey}. Early evaluation protocols compare generated object mentions with annotated visual references, including ground-truth object labels, captions, or object-presence annotations~\citep{rohrbach2018object,li2023evaluating,petryk2024aloha,he2025evaluating}. Other work uses stronger LLMs or LVLMs as external judges to assess response faithfulness~\citep{liu2024mitigating,jing2024faithscore,wang2023evaluation,jiang2024hal,cao2024visdiahalbench}. These methods are effective for controlled evaluation, but they often require annotations, auxiliary models, or multi-stage verification, making them less suitable when object-level references are unavailable.

\paragraph{Training-free internal-signal detectors.}
To reduce dependence on external references, recent work has used signals available during LVLM inference. Likelihood-based methods such as LURE use the negative log-likelihood of generated object tokens as a hallucination cue~\citep{zhou2024analyzing}. Representation-based methods estimate grounding from internal activations. Internal Confidence uses logit-lens probabilities over image hidden states~\citep{jiang2025interpreting}, while Contextual Lens studies contextual embeddings beyond raw logit-lens evidence~\citep{phukan2025beyond}. Attention-based methods examine how object generation relies on image or text tokens. SVAR measures attention to image tokens in intermediate layers, while PAS uses attention to preliminary or generated text tokens to detect prefix-driven hallucinations~\citep{jiang2025devils,hoang2026pas}. GLSim improves object-level detection by combining global and local image-text similarity~\citep{park2025glsim}. 

\paragraph{Confounded support in LVLM hallucination.}
Several studies show that hallucinated objects can receive support from sources other than direct object evidence. Object co-occurrence, instruction frequency, and scene priors can make absent objects appear plausible~\citep{li2023evaluating,zhou2024analyzing}. Mitigation work also suggests that reducing statistical bias or unimodal priors can reduce hallucination, indicating that plausible but unsupported mentions are often prior-driven~\citep{leng2024mitigating}. Language-side effects provide another source of support. LVLMs may show text inertia, producing similar descriptions even when visual evidence is weak~\citep{liu2024paying}, and generated prefix tokens can influence later object mentions~\citep{hoang2026pas}. Image-side signals can also be imperfect. Attention may vary across layers or heads~\citep{jain2019attention,jiang2025devils}, be affected by visual attention sinks~\citep{kang2025see}, or concentrate on background and outlier image tokens rather than object-specific regions~\citep{gong2024damro,che2025hallucinatory}. These findings motivate a two-sided detector that considers both whether an object has image-token evidence and whether its support is more image-driven than prefix-driven.

\section{Method}
\subsection{Problem Setup}
We consider a pretrained LVLM that takes an image \(I\) and a textual instruction \(x\), and autoregressively generates a response
$
y=(y_1,\ldots,y_T).
$
We study object-level hallucination detection. Let \(o\) denote an object mention in the generated response, and let \(t_o\) denote the position of the generated token matched to \(o\). The goal is to assign each object mention a grounding score
$
s(o,I,x,y_{<t_o}) \in \mathbb{R},
$
where larger values indicate stronger visual grounding. A binary detector can be obtained by thresholding
\[
G(o,I,x,y_{<t_o})=
\mathbbm{1}\{s(o,I,x,y_{<t_o})\geq \delta\},
\]
where \(\delta\) is a decision threshold. In our experiments, we evaluate ranking quality using AUROC and AUPR, so no fixed threshold is required.

The LVLM represents the image as a sequence of visual tokens
$
\mathcal{V}=\{v_1,\ldots,v_N\},
$
which are projected into the language-model space. Let \(h_i^\ell\in\mathbb{R}^d\) denote the hidden representation of token \(i\) at layer \(\ell\). We use \(\ell_I\) for image-token representations and \(\ell_T\) for text-token representations. The object-token representation is denoted by
$
h_o^{\ell_T}=h_{t_o}^{\ell_T}.
$
We write \(\mathrm{sim}(\cdot,\cdot)\) for cosine similarity and
$
\mathrm{sim}^{+}(a,b)
$
for its \([0,1]\)-rescaled form.

\paragraph{Visual logit-lens evidence.}
To obtain object-specific evidence from image tokens, we apply logit lens~\citep{nostalgebraist2020logitlens} to image-token hidden states. Let \(W_U\in\mathbb{R}^{d\times |\Omega|}\) be the language-model unembedding matrix, where \(\Omega\) is the vocabulary. For each image token \(v_i\), we compute
$
z_i=h_{v_i}^{\ell_I}W_U$. The visual logit-lens evidence for object \(o\) at image token \(v_i\) is
$
p_i(o)=\mathrm{softmax}(z_i)_o.
$ This gives an object-specific score for each image token.

\subsection{VisER}
VisER assigns each generated object mention a grounding score from two complementary signals. 
Visual Evidence (\textbf{VE}) measures object-specific image-token evidence, while Visual Reliance (\textbf{VR}) compares image-derived support with generated-prefix support. Their combination favors objects that are both visually evidenced and more image-driven than prefix-driven.
\paragraph{Visual Evidence (VE).}

Visual Evidence measures whether object-context compatibility is supported by object-specific evidence from image tokens. Let \(h_{\mathrm{ctx}}^{\ell_T}\) denote the hidden representation of the final prompt token before response generation. We first aggregate the visual logit-lens evidence for object \(o\) over all image tokens:
\[
M_{\mathrm{vis}}(o)=\sum_{i=1}^{N}p_i(o),
\]
where \(p_i(o)\) is the logit-lens probability assigned to object token \(o\) at image token \(v_i\). We then compute an evidence gate
\[
g(o)=
\sigma\left(
\frac{M_{\mathrm{vis}}(o)}{\tau+\epsilon}
\right),
\]
where \(\sigma(\cdot)\) is the sigmoid function, \(\tau\) is an evidence-scale parameter, and \(\epsilon\) is used for numerical stability. We estimate \(\tau\) from an unlabeled calibration split disjoint from the evaluation set, using the average visual evidence mass. The Visual Evidence score is
\[
VE(o)
=
\mathrm{sim}(h_{\mathrm{ctx}}^{\ell_T},h_o^{\ell_T})
\cdot g(o).
\]
This score keeps the object-context compatibility term, but modulates it by object-specific image-token evidence. As a result, weak object-specific image-token evidence reduces the contribution of object-context compatibility to the final VE score.

\paragraph{Visual Reliance (VR).}

Visual Reliance compares image-derived support with support from the generated prefix. This is useful because an object mention may be reinforced by previous tokens even when the image provides limited evidence. We first compute image-derived support as
\[
S_{\mathrm{img}}(o)
=
\sum_{i=1}^{N}
\hat{p}_i(o)\,
\mathrm{sim}^{+}(h_{v_i}^{\ell_I},h_o^{\ell_T}),
\]
where the normalized visual evidence weight is
\[
\hat{p}_i(o)
=
\frac{p_i(o)}
{\sum_{j=1}^{N}p_j(o)+\epsilon}.
\]
Thus, all image tokens contribute to \(S_{\mathrm{img}}(o)\), with larger weight assigned to tokens that provide stronger object-specific evidence.

For prefix support, let \(Y_{<t_o}=\{y_1,\ldots,y_{t_o-1}\}\) be the generated prefix before object \(o\). We use input embeddings as a lightweight estimate of lexical-semantic support from the prefix. Let \(e_o\) be the input embedding of the object token and \(e_j\) the input embedding of prefix token \(y_j\). We select the \(K_o=\min(K,t_o-1)\) prefix tokens most similar to \(e_o\), where \(K\) is the maximum number of selected prefix tokens and \(K_o\) is the actual number of prefix tokens selected for object \(o\). If \(K_o=0\), we set \(S_{\mathrm{text}}(o)=0\). Otherwise,
\[
S_{\mathrm{text}}(o)
=
\frac{1}{K_o}
\sum_{j\in \mathrm{TopK}(Y_{<t_o};o)}
\mathrm{sim}^{+}(e_j,e_o),
\]
where \(\mathrm{TopK}(Y_{<t_o};o)\) returns the selected prefix-token indices. We define Visual Reliance as 
\[
VR(o)
=
\frac{
S_{\mathrm{img}}(o)
}{
S_{\mathrm{img}}(o)+S_{\mathrm{text}}(o)+\epsilon
}.
\]
A larger \(VR(o)\) indicates that support for the object is more image-derived than prefix-derived.

\paragraph{Final VisER Score.}

VE and VR capture complementary aspects of object grounding. \(VE(o)\) down-weights object-context compatibility when object-specific image-token evidence is weak. \(VR(o)\) compares image-derived support with generated-prefix support, reducing the effect of objects that are more strongly supported by the prefix than by the image. We combine the two scores as
\[
s_{\mathrm{VisER}}(o)
=
\alpha VE(o)
+
(1-\alpha)VR(o),
\]
where \(\alpha\in[0,1]\) controls the trade-off between visual evidence and visual reliance. Larger values of \(s_{\mathrm{VisER}}(o)\) indicate stronger visual grounding.

\begin{figure*}[t]
    \centering
    \includegraphics[width=\textwidth]{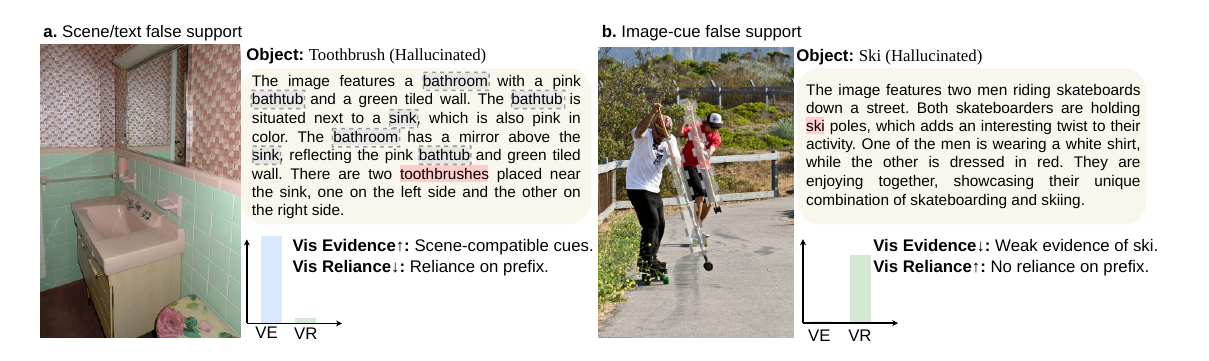}
    \caption{
    Complementary roles of Visual Evidence (VE) and Visual Reliance (VR) in detecting false-support hallucinations. Left: “toothbrushes” is plausible in the bathroom scene and receives relatively high compatibility, but low VR indicates that the mention is not primarily image-driven. Right: ``ski'' receives high VR from image-side cues such as poles and motion, but low VE reveals weak ski-specific evidence. 
    }
    \label{fig:motivation}
\end{figure*}

\paragraph{Why VE and VR are Complementary.} Figure \ref{fig:motivation} shows why both components are necessary. In the bathroom example, the hallucinated object ``toothbrush'' is scene-compatible and can receive high similarity or compatibility because bathrooms, sinks, and toothbrushes commonly co-occur. A similarity-based detector can therefore over-score it; VE alone may also remain high if related visual cues provide weak object-like evidence. However, VR reveals that the mention is not primarily image-driven. In the skateboard example, the hallucinated object ``ski'' is supported by image-side cues such as poles and body motion, so an image-versus-text ratio can be high. However, VE remains low because the image lacks object-specific evidence for an actual ski. These cases show that neither visual support nor visual reliance alone is sufficient; reliable grounding requires their combination.

\subsection{A Bayesian Source-Confounding View}
\label{sec:bayesian_view}

VisER is motivated by the fact that image-conditioned support for an object does not necessarily imply visual grounding. For an object mention \(o\) generated at position \(t_o\), Bayes' rule gives
\[
\begin{aligned}
\log p_\theta(o \mid I,x,y_{<t_o})
&=
\underbrace{\log p_\theta(o \mid x,y_{<t_o})}_{\text{prefix-side support}}
\\
&\quad+
\underbrace{
\log \frac{p_\theta(I \mid o,x,y_{<t_o})}
{p_\theta(I \mid x,y_{<t_o})}
}_{\text{image-side compatibility gain}} .
\end{aligned}
\]
We use this decomposition as an analytical lens; VisER does not estimate probabilities directly. The decomposition exposes two sources of confounding in object hallucination. First, the prefix-side term can make an object likely because it follows naturally from the generated text. Second, the image-side compatibility can be high because an object is scene-plausible, or is associated with visual cues.%, rather than because the object is directly present. 

\paragraph{Prefix-side confounding.}
Visual Reliance addresses prefix-side confounding by asking whether the object receives additional support from the image beyond what the prefix already explains. This mirrors the conditional pointwise mutual-information contrast
\[
\begin{aligned}
\operatorname{PMI}(I;o \mid x,y_{<t_o})
=\\
\log p_\theta(o \mid I,x,y_{<t_o})
-
\log p_\theta(o \mid x,y_{<t_o}),
\end{aligned}
\]
which isolates the gain from conditioning on the image. This contrast suggests that a useful training-free detector should compare image-derived support against prefix-derived support, rather than relying on the image-conditioned object probability alone. Instead of estimating this contrast with additional model calls or explicit likelihoods, VisER constructs internal proxies for the competing sources using \(S_{\mathrm{img}}(o)\) for image-derived support and \(S_{\mathrm{text}}(o)\) for prefix-derived support. Visual Reliance log-odds satisfy
\[
\log \frac{VR(o)}{1-VR(o)}
=
\log S_{\mathrm{img}}(o)
-
\log\!\left(S_{\mathrm{text}}(o)+\epsilon\right).
\]
Thus, \(VR(o)\) is a bounded monotonic transform of a smoothed image-versus-prefix log-ratio. It does not equal the conditional PMI, but it preserves the same source comparison. The score increases when support is more image-derived than prefix-derived, and decreases when the generated prefix better explains the object mention.

\paragraph{Image-side confounding.}
Visual Evidence addresses the remaining image-side confounding. The Bayesian image-side gain measures whether the image is compatible with the hypothesis that \(o\) belongs in the description. The similarity term $\operatorname{sim}(h^{\ell_T}_{ctx},h^{\ell_T}_{o})$ captures whether the object is compatible with the multimodal context, while \(M_{\mathrm{vis}}(o)\) tests whether image tokens provide direct object-specific evidence. In this sense, \(VE(o)\) is a specificity-corrected proxy for the image-side term. It retains compatibility when it is visually evidenced, and suppresses scene-level or cue-driven compatibility that can otherwise mimic grounding.

Thus, a generated object receives a high VisER score only when it is both supported by direct image-token evidence and more strongly explained by the image than by the generated prefix. 

\begin{table*}[!t]
\centering
\resizebox{\textwidth}{!}{
\begin{tabular}{llcccccccc|cc}
\toprule
\multirow{2}{*}{Dataset} & \multirow{2}{*}{Method}
& \multicolumn{2}{c}{LLaVA-1.5-7B}
& \multicolumn{2}{c}{LLaVA-NeXT}
& \multicolumn{2}{c}{InstructBLIP}
& \multicolumn{2}{c}{MiniGPT-4}
& \multicolumn{2}{c}{Average} \\
\cmidrule(lr){3-4}
\cmidrule(lr){5-6}
\cmidrule(lr){7-8}
\cmidrule(lr){9-10}
\cmidrule(lr){11-12}
& & AUROC$\uparrow$ & AUPR$\uparrow$
& AUROC$\uparrow$ & AUPR$\uparrow$
& AUROC$\uparrow$ & AUPR$\uparrow$
& AUROC$\uparrow$ & AUPR$\uparrow$
& AUROC$\uparrow$ & AUPR$\uparrow$ \\
\midrule

\multirow{8}{*}{MSCOCO}
& Entropy \citep{malininuncertainty} 
& 71.41 & 87.66 & 58.25 & 90.09 & 65.81 & 84.25 & 45.78 & 82.44 & 60.31 & 86.11 \\

& NLL \citep{zhou2024analyzing}
& 70.28 & 87.19 & 58.99 & 90.27 & 65.30 & 83.92 & 57.55 & 86.18 & 63.03 & 86.89 \\

& Internal Conf. \citep{jiang2025interpreting}
& 73.88 & 90.62 & 78.00 & 95.83 & 84.59 & 93.57 & 75.79 & 93.50 & 78.07 & 93.38 \\

& SVAR \citep{jiang2025devils}
& 74.39 & 92.20 & 74.02 & 95.45 & 78.19 & 91.95 & 85.94 & 96.63 & 78.14 & 94.06 \\

& Contextual Lens \citep{phukan2025beyond}
& 70.93 & 89.03 & 59.15 & 90.04 & 81.17 & 91.94 & 82.55 & 94.43 & 73.45 & 91.36 \\

& PAS \citep{hoang2026pas}
& 82.52 & 94.81 & 73.71 & 95.58 & 81.96 & 91.58 & 80.96 & 95.16 & 79.79 & 94.28 \\

& GLSim \citep{park2025glsim}
& 84.09 & 94.45 & 77.43 & 95.65 & 82.86 & 93.30 & 80.66 & 94.64 & 81.26 & 94.51 \\ \cmidrule{2-12}
& VisER (Ours)
& \textbf{86.17} & \textbf{95.37}
& \textbf{81.44} & \textbf{96.74}
& \textbf{85.66} & \textbf{94.84}
& \textbf{86.29} & \textbf{96.35}
& \textbf{84.89} & \textbf{95.82} \\

\midrule

\multirow{8}{*}{Pascal VOC}
% \multirow{8}{*}{Pascal VOC}
& Entropy \citep{malininuncertainty}
& 53.12 & 74.34 & 62.39 & 91.07 & 60.57 & 76.11 & 53.48 & 81.33 & 57.39 & 80.71 \\

& NLL \citep{zhou2024analyzing}
& 52.56 & 74.24 & 62.57 & 91.49 & 60.44 & 76.62 & 57.97 & 83.47 & 58.39 & 81.46 \\

& Internal Conf. \citep{jiang2025interpreting}
& 71.05 & 82.74 & 77.50 & 95.47 & 82.96 & 89.96 & 68.58 & 86.15 & 75.02 & 88.58 \\

& SVAR \citep{jiang2025devils}
& 74.59 & 88.75 & 73.84 & 95.11 & 75.86 & 87.51 & \textbf{82.62} & \textbf{94.74} & 76.73 & 91.53 \\

& Contextual Lens \citep{phukan2025beyond}
& 70.73 & 84.82 & 58.87 & 90.12 & 80.62 & 88.87 & 74.69 & 88.80 & 71.23 & 88.15 \\

& PAS \citep{hoang2026pas}
& 80.31 & 91.36 & 74.96 & 95.66 & 79.30 & 86.85 & 78.20 & 92.80 & 78.19 & 91.67 \\

& GLSim \citep{park2025glsim}
& 80.27 & 90.58 & 79.17 & 95.88 & 84.89 & 92.49 & 78.72 & 93.41 & 80.76 & 93.09 \\ \cmidrule{2-12}
& VisER (Ours)
& \textbf{82.20} & \textbf{91.26}
& \textbf{87.31} & \textbf{98.08}
& \textbf{85.72} & \textbf{93.24}
& 80.34 & 92.93
& \textbf{83.89} & \textbf{93.88} \\
\bottomrule
\end{tabular}
}
\caption{Main object-level hallucination detection results on MSCOCO and Pascal VOC. AUROC and AUPR are reported for each LVLM setting, with averages computed across the four model settings.
}
\label{tab:main_results}
\end{table*}

\begin{table*}[!t]
\centering

\resizebox{\textwidth}{!}{
\begin{tabular}{llcccccccc|cc}
\toprule
\multirow{2}{*}{Dataset} & \multirow{2}{*}{Method}
& \multicolumn{2}{c}{LLaVA-1.5-13B}
& \multicolumn{2}{c}{InternVL3}
& \multicolumn{2}{c}{Shikra}
& \multicolumn{2}{c}{Qwen2.5-VL}
& \multicolumn{2}{c}{Average} \\
\cmidrule(lr){3-4}
\cmidrule(lr){5-6}
\cmidrule(lr){7-8}
\cmidrule(lr){9-10}
\cmidrule(lr){11-12}
& & AUROC$\uparrow$ & AUPR$\uparrow$
& AUROC$\uparrow$ & AUPR$\uparrow$
& AUROC$\uparrow$ & AUPR$\uparrow$
& AUROC$\uparrow$ & AUPR$\uparrow$
& AUROC$\uparrow$ & AUPR$\uparrow$ \\
\midrule

\multirow{8}{*}{MSCOCO}
& Entropy                         & 70.99 & 88.85 & 64.38 & 95.33 & 69.20 & 87.62 & 68.10 & 95.85 & 68.17 & 91.91 \\
& NLL                             & 69.11 & 88.10 & 64.30 & 95.31 & 66.66 & 86.93 & 65.53 & 95.50 & 66.40 & 91.46 \\
& Internal Conf.                  & 70.30 & 89.41 & 66.63 & 95.56 & 68.53 & 89.23 & 64.80 & 95.20 & 67.56 & 92.35 \\
& SVAR                            & 72.83 & 92.63 & 74.02 & 96.98 & 59.73 & 84.76 & 73.76 & 96.76 & 70.08 & 92.78 \\
& Contextual Lens                 & 76.39 & 92.31 & 77.18 & 97.15 & 78.72 & 91.80 & 71.25 & 96.17 & 75.88 & 94.36 \\
& PAS                             & 80.79 & 94.76 & 75.44 & 97.41 & 79.48 & 93.56 & 73.95 & 97.18 & 77.42 & 95.73 \\
& GLSim                           & 81.53 & 94.82 & 67.23 & 95.79 & 75.03 & 91.74 & 69.67 & 96.11 & 73.36 & 94.62 \\ \cmidrule{2-12}
& VisER                           & \textbf{83.82} & \textbf{95.35} & \textbf{82.54} & \textbf{97.78} & \textbf{80.18} & \textbf{93.68} & \textbf{80.10} & \textbf{97.30} & \textbf{81.66} & \textbf{96.03} \\
\bottomrule
\end{tabular}
}
\caption{Extended MSCOCO results on additional LVLMs. AUROC and AUPR are reported for each model, with averages computed across the four extended settings.}
\label{tab:extended_results}
\end{table*}

\section{Experiments}

\paragraph{Models.}
We evaluate VisER on a diverse set of LVLMs covering different model families and scales. Our main evaluation includes LLaVA-1.5-7B~\citep{liu2023visual}, LLaVA-NeXT-7B~\citep{liu2024improved}, InstructBLIP~\citep{dai2023instructblip}, and MiniGPT-4~\citep{zhu2024minigpt}. We include LLaVA-NeXT-7B to evaluate generalization to a more recent LLaVA-family architecture. To further examine robustness across architectures, we also report results on LLaVA-1.5-13B, InternVL3-8B~\citep{zhu2025internvl3}, Shikra-7B~\citep{chen2023shikra}, and Qwen2.5-VL \citep{bai2025qwen3}. All models use greedy decoding with a maximum output length of 512 new tokens. Implementation details, including model configurations, layer selection, and hyperparameter settings, are provided in Appendix~\ref{sec:appendix}. \footnote{The code is publicly available in \url{https://github.com/AfsanehEB/VisER}.}

\paragraph{Benchmarks.}
We evaluate on MSCOCO~\citep{lin2014microsoft}, following prior object hallucination detection protocols~\citep{hoang2026pas,jiang2025devils}, and additionally report results on Pascal VOC~\citep{everingham2010pascal}. Following~\citep{li2023evaluating, leng2024mitigating, jiang2025interpreting} we randomly sample 500 images from the validation split. MSCOCO and Pascal VOC contain 80 and 20 object categories, respectively, with image-level object annotations indicating which categories are present in each image. Following the CHAIR object-level evaluation protocol, we extract generated object mentions and label them as grounded or hallucinated by matching them against the annotated object lists. All LVLMs are prompted with: ``Describe the given image in detail.''

\paragraph{Baselines.}
We compare VisER with several training-free object hallucination detection baselines. These include logit-based uncertainty signals, namely Negative Log-Likelihood (NLL)~\citep{zhou2024analyzing} and Entropy~\citep{malininuncertainty}; attention-based methods, including Summed Visual Attention Ratio (SVAR)~\citep{jiang2025devils} and Prelim Attention Score (PAS)~\citep{hoang2026pas}; and representation-based methods, including Internal Confidence (IC)~\citep{jiang2025interpreting} and Global-Local Similarity (GLSim)~\citep{park2025glsim}. All baselines are evaluated on the same generated captions, object mentions, and image samples for each benchmark to ensure a fair comparison. The baseline details can be found in Appendix~\ref{app:baseline}. %All scores are oriented so that larger values indicate stronger visual grounding. 
We report object-level AUROC and AUPR, using grounded object mentions as the positive class. For parameterized baselines, we adopt the hyperparameter settings specified by the original methods when available. %To compare against prompting-based object verification, we additionally evaluate POPE-style \citep{li2023evaluating} yes/no verification on the generated object mentions, as described in Section~\ref{sec:pope}.

\subsection{Results}
\paragraph{Main results.}
Table~\ref{tab:main_results} reports object-level hallucination detection results on MSCOCO and Pascal VOC across four LVLM settings. VisER achieves the strongest average performance on both datasets. On MSCOCO, VisER obtains the best AUROC for all four models, with an average AUROC of 84.89, improving over the strongest average baseline, GLSim, by 3.63 points. It also achieves the highest average AUPR of 95.82. On Pascal VOC, VisER again achieves the best average performance, reaching 83.89 AUROC and 93.88 AUPR.

The gains are especially pronounced on LLaVA-NeXT and InstructBLIP. On MSCOCO, VisER improves LLaVA-NeXT AUROC from the strongest baseline score of 78.00 to 81.44. On Pascal VOC, it improves over the strongest baseline by 8.14 points on LLaVA-NeXT and 0.83 points on InstructBLIP. This shows that hallucinated objects can receive strong internal support from scene priors or autoregressive prefix cues, and that combining object-specific visual evidence with image-versus-prefix reliance improves detection.

\paragraph{Extended models.}
Table~\ref{tab:extended_results} reports additional MSCOCO results on LLaVA-1.5-13B, InternVL3, Shikra-7B, and Qwen2.5-VL. VisER achieves the best average AUROC of 81.66, improving over the strongest average baseline, PAS, by 4.24 points. The gains are largest on Qwen2.5-VL and InternVL3, where VisER improves AUROC over the strongest baselines by 6.15 and 5.36 points, respectively. VisER also achieves the best results on LLaVA-1.5-13B and Shikra-7B, showing that the two-sided grounding view generalizes across model scales and architectures.

\begin{table}[t]
\centering
\small
\setlength{\tabcolsep}{3.2pt}
\begin{tabular}{llrrrrrr}
\toprule
Model & Method & Time & ACC & PR(T) & PR(F) & F1 & F1(F) \\
\midrule
\multirow{2}{*}{7B}
& POPE & 3.19 & 67.77 & 61.28 & 91.86 & 74.97 & 54.75 \\
& VisER & 2.36 & 78.67 & 80.93 & 76.72 & 77.86 & 79.42 \\
\midrule
\multirow{2}{*}{13B}
& POPE & 4.50 & 60.53 & 56.05 & 90.72 & 71.20 & 37.29 \\
& VisER & 3.27 & 76.56 & 76.36 & 76.78 & 76.65 & 76.47 \\
\bottomrule
\end{tabular}
\caption{
Balanced object-level comparison with POPE. Time denotes object verification overhead.%; VisER does not require additional object-level generations.
}
\label{tab:balanced_pope}
\end{table}

\begin{table}[t]
\centering
\small
\setlength{\tabcolsep}{5pt}
\begin{tabular}{lccc}
\toprule
Method & Metric & Peak VRAM (GB) \\
\midrule
Entropy/NLL & Token Probability & 13.33 \\
Internal Conf. & Token Probability & 15.77  \\
GLSim &  Embedding-similarity & 13.54  \\
SVAR/PAS & Attention Allocation & 14.31  \\
VisER & Embedding-based & 13.33 \\
\bottomrule
\end{tabular}
\caption{
Post-caption detector-stage profiling on LLaVA-1.5-7B in float16 mode with batch size 1.
}
\label{tab:detector_memory}
\end{table}
\subsection{Comparison with POPE}
\label{sec:pope}

\begin{figure*}[t]
\centering
\begin{tikzpicture}

% ---------------------------------------------------------
% (a) Effect of K
% ---------------------------------------------------------
\begin{axis}[
    name=plotK,
    width=0.34\textwidth,
    height=0.25\textwidth,
    xlabel={$K$},
    ylabel={AUROC},
    xmin=2, xmax=20,
    ymin=0.80, ymax=0.865,
    xtick={2,4,6,8,10,12,14,16,18,20},
    ytick={0.82,0.83,0.84,0.85,0.86, 0.87},
    grid=both,
    grid style={line width=.1pt, draw=gray!20},
    major grid style={line width=.2pt, draw=gray!35},
    tick label style={font=\scriptsize},
    label style={font=\small},
    title style={font=\small},
    legend style={
        font=\scriptsize,
        at={(0.98,0.10)},
        anchor=south east,
        draw=none,
        fill=none
    }
]
\addplot[
    mark=*,
    mark size=0.6pt,
    thick,
    orange
] coordinates {
    (2,0.8485)
    (4,0.8552)
    (6,0.8589)
    (8,0.8607)
    (10,0.8617)
    (12,0.8624)
    (14,0.8628)
    (16,0.8629)
    (18,0.8629)
    (20,0.8629)
};
\addlegendentry{VisER}

\addplot[
    dashed,
    thick,
    green!70!black
] coordinates {
    (2,0.8079)
    (20,0.8079)
};
\addlegendentry{PAS}

\addplot[
    dashed,
    thick,
    blue
] coordinates {
    (2,0.8409)
    (20,0.8409)
};
\addlegendentry{GLSim}

\end{axis}

% ---------------------------------------------------------
% (b) Effect of alpha
% ---------------------------------------------------------
\begin{axis}[
    name=plotAlpha,
    at={(plotK.east)},
    anchor=west,
    xshift=1.40cm,
    width=0.34\textwidth,
    height=0.25\textwidth,
    xlabel={$\alpha$},
    ylabel={AUROC},
    xmin=0, xmax=1,
    ymin=0.77, ymax=0.865,
    xtick={0,0.2,0.4,0.6,0.8,1.0},
    ytick={0.78,0.80,0.82,0.84,0.86},
    grid=both,
    grid style={line width=.1pt, draw=gray!20},
    major grid style={line width=.2pt, draw=gray!35},
    tick label style={font=\scriptsize},
    label style={font=\small},
    title style={font=\small},
    legend style={
        font=\scriptsize,
        at={(0.98,0.10)},
        anchor=south east,
        draw=none,
        fill=none
    }
]
\addplot[
    mark=*,
    mark size=0.6pt,
    thick,
    orange
] coordinates {
    (0.00,0.7822)
    (0.05,0.8077)
    (0.10,0.8256)
    (0.15,0.8387)
    (0.20,0.8477)
    (0.25,0.8543)
    (0.30,0.8585)
    (0.35,0.8609)
    (0.40,0.8617)
    (0.45,0.8614)
    (0.50,0.8601)
    (0.55,0.8584)
    (0.60,0.8565)
    (0.65,0.8543)
    (0.70,0.8519)
    (0.75,0.8496)
    (0.80,0.8472)
    (0.85,0.8451)
    (0.90,0.8428)
    (0.95,0.8408)
    (1.00,0.8388)
};
\addlegendentry{VisER}

\addplot[
    dashed,
    thick,
    olive
] coordinates {
    (0,0.8388)
    (1,0.8388)
};
\addlegendentry{$VE(o)$}

\addplot[
    dashed,
    thick,
    cyan
] coordinates {
    (0,0.7822)
    (1,0.7822)
};
\addlegendentry{$VR(o)$}

\end{axis}

% ---------------------------------------------------------
% (c) Effect of image layer
% ---------------------------------------------------------
\begin{axis}[
    name=plotLayer,
    at={(plotAlpha.east)},
    anchor=west,
    xshift=1.40cm,
    width=0.34\textwidth,
    height=0.25\textwidth,
    xlabel={Image layer},
    ylabel={AUROC},
    xmin=0, xmax=32,
    ymin=0.73, ymax=0.87,
    xtick={0,4,8,12,16,20,24,28,32},
    ytick={0.74, 0.76,0.78,0.80,0.82,0.84,0.86},
    grid=both,
    grid style={line width=.1pt, draw=gray!20},
    major grid style={line width=.2pt, draw=gray!35},
    tick label style={font=\scriptsize},
    label style={font=\small},
    title style={font=\small},
]
\addplot[
    mark=*,
    mark size=0.6pt,
    thick,
    orange
] coordinates {
    (0,0.7507)
    (1,0.7482)
    (2,0.7495)
    (3,0.7494)
    (4,0.7502)
    (5,0.7546)
    (6,0.7570)
    (7,0.7601)
    (8,0.7591)
    (9,0.7552)
    (10,0.7523)
    (11,0.7482)
    (12,0.7513)
    (13,0.7502)
    (14,0.7451)
    (15,0.7497)
    (16,0.7515)
    (17,0.7568)
    (18,0.7658)
    (19,0.7974)
    (20,0.8019)
    (21,0.8029)
    (22,0.8196)
    (23,0.8245)
    (24,0.8271)
    (25,0.8386)
    (26,0.8533)
    (27,0.8575)
    (28,0.8615)
    (29,0.8537)
    (30,0.8527)
    (31,0.8345)
    (32,0.8601)
};
\end{axis}

\node at (plotK.south) [yshift=-30pt] {\small (a)};
\node at (plotAlpha.south) [yshift=-30pt] {\small (b)};
\node at (plotLayer.south) [yshift=-30pt] {\small (c)};

\end{tikzpicture}
\caption{
Sensitivity analysis of VisER.
Left: effect of the number of selected prefix tokens $K$.
Middle: effect of the interpolation weight $\alpha$.
Right: effect of the image layer when the text layer is fixed.
}
\label{fig:sensitivity_k_alpha_layer}
\end{figure*}
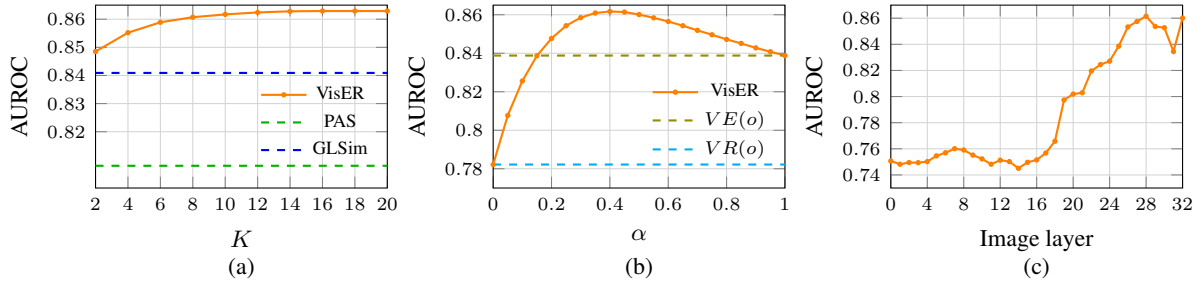

We further compare VisER with POPE-style prompting-based object verification~\citep{li2023evaluating}. Given the objects extracted from each generated caption, POPE verifies each object with the question:
\texttt{Q: Is there a \{object\} in the image?}
%Each object is labeled as grounded or hallucinated using the same MSCOCO object-matching protocol as in our main evaluation. 
For POPE, we use yes/no parsing, and an object is predicted as present only if the response starts with an affirmative token such as ``yes''.

For this POPE comparison, we use a larger 3000-image MSCOCO subset to obtain a sufficient number of hallucinated object mentions. Because generated object sets are naturally imbalanced, we construct balanced object-level subsets by retaining all hallucinated objects and randomly sampling an equal number of grounded objects. This yields 2495 grounded and 2495 hallucinated objects for LLaVA-1.5-7B and 2216 grounded and 2216 hallucinated for LLaVA-13B. We repeat the sampling procedure 1000 times and report the average.% VisER thresholds are fixed before balancing and are not re-tuned on the balanced subsets.

As shown in Table~\ref{tab:balanced_pope}, VisER improves ACC from 67.77 to 78.67 and hallucinated-object F1 from 54.75 to 79.42, while reducing verification time from 3.19s to 2.36s. A similar pattern is observed for LLaVA-13B, indicating that these gains also hold at a larger model scale. POPE achieves high precision for grounded objects, but its low F1(F) indicates a bias toward predicting object presence. In contrast, VisER provides stronger hallucination-sensitive discrimination without requiring additional yes/no generations.% for each candidate object.

\paragraph{Efficiency.}
Finally, Table~\ref{tab:detector_memory} reports the post-caption detector-stage peak GPU memory on LLaVA-1.5-7B. VisER matches the lightweight Entropy/NLL baseline in peak VRAM, while requiring less memory than Internal Confidence, GLSim, and attention-based scoring. This indicates that VisER can be implemented efficiently by retaining only selected hidden representations and computing logit-lens evidence in chunks. All VRAM measurements are reported on a single NVIDIA A100 GPU using the same inference setting.

\subsection{Ablation Study}
\label{main:ablation}
\paragraph{Sensitivity Analysis.}Figure~\ref{fig:sensitivity_k_alpha_layer} studies the sensitivity of VisER to \(K\), \(\alpha\), and the image layer on LLaVA-1.5-7B using the MSCOCO dataset. Figure~\ref{fig:sensitivity_k_alpha_layer}(a) shows that VisER remains above the strongest baselines, PAS and GLSim, across the full range of \(K\). Performance improves from 84.85 AUROC at \(K=2\) to 85.66 at \(K=10\), and then saturates, indicating that a small set of highly related prefix tokens is sufficient to estimate text-prefix support. Figure~\ref{fig:sensitivity_k_alpha_layer}(b) shows that both components contribute to performance. \(VR(o)\) alone obtains 78.22 AUROC and \(VE(o)\) alone reaches 83.88, while their combination performs best, peaking at 85.66 around \(\alpha=0.4\). This confirms that visual evidence and visual reliance capture complementary hallucination cues. Figure~\ref{fig:sensitivity_k_alpha_layer}(c) shows that later image layers provide stronger evidence, with the best performance at layer 32, suggesting that semantically richer image-token representations improve object-level grounding signals.

\begin{figure}[t]
\centering
\begin{tikzpicture}
\begin{axis}[
    width=1.08\columnwidth,
    height=0.5\columnwidth,
    ybar,
    bar width=6pt,
    ymin=60,
    ymax=86,
    ytick={60,65,70,75,80,85},
    ylabel={AUROC (\%)},
    ylabel style={font=\small},
    xmin=0.4,
    xmax=7.5,
    xtick={1,2,3,4,5,6},
    xticklabels={LLaVA-7B,LLaVA-13B,MiniGPT-4,LLaVA-NeXT,InternVL3,Shikra},
    xticklabel style={font=\scriptsize, rotate=25, anchor=east},
    yticklabel style={font=\scriptsize},
    enlarge x limits=false,
    axis line style={black},
    tick style={black},
    legend style={
        at={(1.003,0.98)},
        anchor=north east,
        draw=gray!25,
        fill=white,
        font=\scriptsize,
        cells={anchor=west}
    },
    legend image code/.code={
        \draw[#1, fill=#1] (0cm,-0.08cm) rectangle (0.30cm,0.08cm);
    },
]

% global_orig
\addplot[
    fill={rgb,255:red,57;green,68;blue,122},
    draw=none
] coordinates {
    (1,72.25)
    (2,65.15)
    (3,71.45)
    (4,72.43)
    (5,67.71)
    (6,67.54)
};

% global_eg
\addplot[
    fill={rgb,255:red,174;green,194;blue,234},
    draw=none
] coordinates {
    (1,83.88)
    (2,80.01)
    (3,81.06)
    (4,79.64)
    (5,81.90)
    (6,72.65)
};

\legend{W/O Gating, W Gating }
\end{axis}
\end{tikzpicture}
\caption{Effect of evidence gating in Visual Evidence. ``W/O Gating'' uses only object-context compatibility, while ``W Gating'' multiplies compatibility by the evidence gate \(g(o)\). %Gating improves AUROC across LVLMs, showing that compatibility is more reliable when validated by object-specific image-token evidence.
}
\label{fig:gating_ablation}
\end{figure}
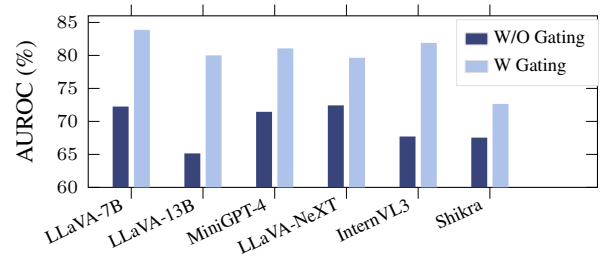

\paragraph{Effect of Evidence Gating.}
Figure~\ref{fig:gating_ablation} isolates the role of the evidence gate \(g(o)\) in Visual Evidence across LVLMs. Without the gate, VE relies only on object-context compatibility and can over-score hallucinations that are plausible from scene context alone. Gating filters compatibility through object-specific image-token evidence, yielding consistent AUROC gains. This supports our central claim that compatibility is a useful but incomplete grounding cue unless its visual source is validated.

\paragraph{Counterfactual Validation of Source Attribution.}
To further examine whether VE and VR capture the intended sources of support, we perform three counterfactual interventions on LLaVA-1.5-7B and LLaVA-1.5-13B: blank-image replacement, patch shuffling, and prefix removal. Table~\ref{tab:counterfactual} reports the corresponding AUROC values.

\begin{table}[t]
\centering
\small
\setlength{\tabcolsep}{3pt}
\begin{tabular}{llccc}
\toprule
Model & Condition & VE & VR & VisER \\
\midrule
\multirow{4}{*}{LLaVA-1.5-7B}
& Original & 83.88 & 78.22 & 86.17 \\
& Blank image & 67.74 & 70.86 & 71.38 \\
& Patch-shuffled & 74.34 & 74.44 & 77.99 \\
& Prefix removed & 83.88 & 50.00 & 83.88 \\
\midrule
\multirow{4}{*}{LLaVA-1.5-13B}
& Original & 80.01 & 77.78 & 83.82 \\
& Blank image & 55.58 & 68.34 & 61.89 \\
& Patch-shuffled & 69.60 & 73.89 & 74.93 \\
& Prefix removed & 80.01 & 50.00 & 80.01 \\
\bottomrule
\end{tabular}
\caption{Counterfactual validation of Visual Evidence and Visual Reliance. Values are AUROC.}
\label{tab:counterfactual}
\end{table}

As shown in Table~\ref{tab:counterfactual}, replacing the image with a blank input or shuffling its patches reduces VE and the full VisER score on both model scales. This suggests that these signals depend on coherent visual evidence rather than only on the presence of image tokens. In contrast, removing prefix support leaves VE unchanged while reducing VR to chance-level discrimination. These results provide additional empirical support for interpreting VE as visual-evidence-sensitive and VR as measuring the balance between image- and prefix-derived support, without claiming calibrated causal attribution.

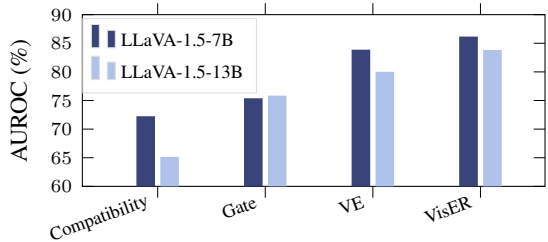
\begin{figure}[t]
\centering
\begin{tikzpicture}
\begin{axis}[
    width=\columnwidth,
    height=0.50\columnwidth,
    ybar,
    bar width=7pt,
    ymin=60,
    ymax=90,
    ytick={60,65,70,75,80,85,90},
    ylabel={AUROC (\%)},
    ylabel style={font=\small},
    xmin=0.3,
    xmax=4.6,
    xtick={1,2,3,4},
    xticklabels={Compatibility,Gate,VE,VisER},
    xticklabel style={
        font=\scriptsize,
        rotate=20,
        anchor=east
    },
    yticklabel style={font=\scriptsize},
    axis line style={black},
    tick style={black},
    legend style={
    at={(0.00,0.98)},
    anchor=north west,
    draw=gray!25,
    fill=white,
    font=\scriptsize,
    cells={anchor=west}
    }
]
\addplot[
    fill={rgb,255:red,57;green,68;blue,122},
    draw=none
] coordinates {
    (1,72.25)
    (2,75.39)
    (3,83.88)
    (4,86.17)
};
\addplot[
    fill={rgb,255:red,174;green,194;blue,234},
    draw=none
] coordinates {
    (1,65.15)
    (2,75.85)
    (3,80.01)
    (4,83.82)
};
\legend{LLaVA-1.5-7B,LLaVA-1.5-13B}
\end{axis}
\end{tikzpicture}
\caption{
Component ablation on MSCOCO. Compatibility uses object-context compatibility alone, Gate uses the evidence gate alone, VE combines compatibility with the evidence gate, and VisER further incorporates Visual Reliance.
}
\label{fig:component_ablation}
\end{figure}

\paragraph{Component Ablation.}
To further isolate the contribution of each component, we compare object-context compatibility alone, the evidence gate alone, Visual Evidence, and the full VisER score. Figure~\ref{fig:component_ablation} reports the corresponding AUROC values.

As shown in Figure~\ref{fig:component_ablation}, the evidence gate is informative on its own, but performs better when used to validate object-context compatibility. Combining compatibility with the evidence gate increases AUROC from 72.25 to 83.88 on LLaVA-1.5-7B and from 65.15 to 80.01 on LLaVA-1.5-13B. The full VisER score further improves performance to 86.17 and 83.82 AUROC, respectively. This supports using the gate as an evidence validator and further shows that VE and VR provide complementary information.

\section{Conclusion}
We introduced VisER, a training-free metric for object-level hallucination detection in LVLMs. VisER addresses source confounding by combining Visual Evidence, measuring whether object-context compatibility is supported by object-specific image-token evidence, with Visual Reliance, comparing image-derived support against generated-prefix support. The object-level VisER score could also support downstream applications such as selective abstention, caption revision, or future decoding-time intervention. Across multiple LVLMs and datasets, VisER consistently improves AUROC and AUPR over strong baselines. Ablations show that the two signals are complementary, making VisER an effective post-hoc detector for visually unsupported object mentions in open-ended generations. 

\section*{Limitations}
VisER focuses on object-existence hallucinations in open-ended image descriptions. Our evaluation primarily follows COCO/VOC-style object categories and therefore does not directly assess open-vocabulary hallucinations, attributes, relations, counts, actions, or other fine-grained factual errors. The object-token representation may also be sensitive to multi-token expressions, subword tokenization, and repeated object mentions. VisER requires access to internal LVLM representations, making it more suitable for white-box or gray-box settings than fully black-box APIs. Future work could extend the two-sided grounding view to phrase- and region-level hallucination detection, and explore whether VE- and VR-style signals can guide hallucination mitigation during decoding. VisER is intended as a diagnostic object-level hallucination detector and should not be treated as a guarantee of visual factuality in safety-critical deployments.

\section*{Acknowledgements}
This research was supported by The University of Melbourne’s Research Computing Services, the Petascale Campus Initiative, and the Spartan HPC facilities. This Spartan facility was established with the assistance of LIEF Grant LE170100200. Moreover, this research was supported by the ARC Centre of Excellence for Automated Decision-Making and Society (CE200100005), and partially funded by the Australian Government through the Australian Research Council.

\bibliography{custom}

\appendix
\section*{Appendix}
\section{Related Work: Baseline Scoring Functions}
\label{app:baseline}
For all baselines, we compute an object-level score for a generated object mention \(o=y_t\), where \(t\) denotes the decoding position of the matched object token. All scores are oriented such that larger values indicate stronger visual grounding and smaller values indicate a higher likelihood of hallucination.

\paragraph{Entropy.}
The entropy baseline measures the uncertainty of the LVLM next-token distribution at the object generation step. Let
\[
q_t(y)=p_\theta(y\mid I,x,y_{<t})
\]
denote the probability assigned to vocabulary token \(y\in\Omega\) when generating \(y_t\). The token-level entropy is
\[
H(o)
=
-
\sum_{y\in\Omega}
q_t(y)\log q_t(y).
\]
Since lower entropy indicates higher model confidence, we use the negative entropy as the grounding-oriented score:
\[
S_{\mathrm{Ent}}(o)=-H(o).
\]

\paragraph{Internal Confidence.}
Internal Confidence estimates object evidence from visual-token representations using logit-lens probabilities. For an image token \(v_i\) at layer \(\ell\), let
\[
p_{\ell,i}(o)
=
\mathrm{softmax}\!\left(h_{v_i}^{\ell}W_U\right)_o
\]
denote the probability assigned to the object token \(o\) after projecting the visual-token hidden state through the language-model unembedding matrix \(W_U\). The Internal Confidence score is defined as the strongest object evidence over image tokens and layers:
\[
S_{\mathrm{IC}}(o)
=
\max_{\ell\in\mathcal{L}}
\max_{i\in\mathcal{I}}
p_{\ell,i}(o),
\]
where \(\mathcal{I}\) denotes the set of image-token positions and \(\mathcal{L}\) is the set of evaluated layers.

\paragraph{SVAR.}
SVAR measures how much attention the generated object token assigns to image tokens. Let \(A_{\ell,h}(t,i)\) denote the attention weight from the generated object position \(t\) to image-token position \(i\), at layer \(\ell\) and head \(h\). We compute the visual attention ratio over the selected middle layers as
\[
S_{\mathrm{SVAR}}(o)
=
\frac{1}{|\mathcal{L}|H}
\sum_{\ell\in\mathcal{L}}
\sum_{h=1}^{H}
\sum_{i\in\mathcal{I}}
A_{\ell,h}(t,i),
\]
where \(H\) is the number of attention heads. Following the original setting, we use the middle-layer range \(\mathcal{L}=\{5,\ldots,18\}\). Larger values indicate stronger visual-token attention during object generation.

\paragraph{Contextual Lens.}
Contextual Lens uses representation-level similarity between the generated object token and visual-token representations. Given the object hidden state \(h_o^{\ell_T}\) and image-token hidden states \(\{h_{v_i}^{\ell_I}\}_{i\in\mathcal{I}}\), we compute
\[
S_{\mathrm{CL}}(o)
=
\max_{i\in\mathcal{I}}
\mathrm{sim}
\left(
h_o^{\ell_T},
h_{v_i}^{\ell_I}
\right).
\]

This score captures the strongest visual-token representation that is compatible with the generated object mention.

\paragraph{GLSim.}
GLSim measures object grounding through global and local image-text compatibility. 
\[
S_{\mathrm{GLSim}}(o)
=
\beta S_{\mathrm{global}}(o)
+
(1-\beta) S_{\mathrm{local}}(o).
\]
The global component measures compatibility between the generated object representation and the multimodal context representation:
\[
S_{\mathrm{global}}(o)
=
\mathrm{sim}
\left(
h_{\mathrm{ctx}}^{\ell_T},
h_o^{\ell_T}
\right),
\]
For the local component, it first selects the \(K\) image tokens with the highest logit-lens evidence for the object:
\[
\mathcal{K}_{o}^{\mathrm{LL}}
=
\mathrm{TopK}_{i\in\mathcal{I}}
\; p_i(o),
\]
where
\[
p_i(o)=\mathrm{softmax}(h_{v_i}^{\ell_I}W_U)_o.
\]
The local score is then computed as the average similarity between the selected image-token representations and the object-token representation:
\[
S_{\mathrm{local}}(o)
=
\frac{1}{K}
\sum_{i\in\mathcal{K}_{o}^{\mathrm{LL}}}
\mathrm{sim}
\left(
h_{v_i}^{\ell_I},
h_o^{\ell_T}
\right).
\]
This baseline measures whether the generated object is compatible with the global and local visual representations.

\paragraph{PAS.}
PAS estimates attention to see whether object generation is driven by previously generated text tokens. Let \(\mathcal{P}_{<t}\) denote the set of generated prefix-token positions before the object token \(o=y_t\). The preliminary attention score is
\[
T_{\mathrm{PAS}}(o)
=
\frac{1}{|\mathcal{L}|H}
\sum_{\ell\in\mathcal{L}}
\sum_{h=1}^{H}
\sum_{j\in\mathcal{P}_{<t}}
A_{\ell,h}(t,j).
\]
A larger \(T_{\mathrm{PAS}}(o)\) indicates that the object token attends more strongly to previously generated text, suggesting that the mention may be text-prefix-driven rather than visually grounded. To keep all baselines oriented consistently, we use
\[
S_{\mathrm{PAS}}(o)
=
-
T_{\mathrm{PAS}}(o).
\]
Thus, larger \(S_{\mathrm{PAS}}(o)\) corresponds to stronger grounding-oriented evidence.

\section{Implementation Details}
\label{sec:appendix}
We implement VisER as a post-hoc object-level hallucination detector applied to generated captions. Each model generates at most 512 new tokens. Greedy decoding is used as the default generation setting in the main experiments for determinism and comparability with prior work. In ablation studies in Appendix \ref{app:ablation}, we additionally evaluate robustness to other decoding strategies. VisER is then computed on the generated object mentions and does not modify the decoding process. Following prior LVLM object hallucination evaluations \citep{jiang2025interpreting, li2023evaluating, phukan2025beyond}, we evaluate on 500 randomly sampled MSCOCO validation images. We use a small held-out calibration split for selecting $\tau$ and report sensitivity analyses for $K$, $\alpha$, and layer choice in Ablation studies in Section \ref{main:ablation} to show that the method is not narrowly tuned to a single configuration. The evidence-scale parameter \(\tau\) is estimated from a disjoint calibration split of 100 images by averaging the visual evidence mass \(M_{\mathrm{vis}}\). The layer indices \((\ell_I,\ell_T)\), the number of selected prefix tokens \(K\), and the final weighting parameter \(\alpha\) are fixed per model, as shown in Table~\ref{tab:hyperparams}. 
For object words that tokenize into multiple subword units, we use the first matched token to compute the scores. For each generated response, if the same object word appears multiple times, we score only its first generated occurrence, since the first token
often captures the core semantic meaning of the object \cite{park2025glsim}. Table \ref{tab:object_counts} reports the number of grounded and hallucinated object mentions.

\begin{table}[!htbp]
\centering
\small
\begin{tabular}{lccc}
\toprule
Model & $(\ell_I,\ell_T)$ & $K$ & $\alpha$ \\
\midrule
LLaVA-1.5-7B             & $(32,31)$ & 10 & 0.40 \\
LLaVA-1.5-13B            & $(40,38)$ & 10 & 0.40 \\
LLaVA-NeXT-7B            & $(32,31)$ & 10 & 0.50 \\
InstructBLIP   & $(32,31)$ & 10 & 0.40 \\
MiniGPT-4      & $(32,27)$ & 10 & 0.25 \\
Shikra-7B                & $(32,27)$ & 10 & 0.20 \\
\bottomrule
\end{tabular}
\caption{
Hyperparameters used for VisER. \(\ell_I\) denotes the image-token layer used for visual evidence and image-derived support, \(\ell_T\) denotes the text/object representation layer, \(K\) is the number of top prefix tokens used for text-prefix support, and \(\alpha\) balances Visual Evidence and Visual Reliance.
}
\label{tab:hyperparams}
\end{table}
\begin{table}[!htbp]
\centering
\small
\begin{tabular}{lccc}
\toprule
Model & Grounded & Hallucinated \\
\midrule
LLaVA-1.5-7B             & 1510 & 416\\
LLaVA-1.5-13B            & 1560 & 373\\
LLaVA-NeXT-7B            & 1173 & 157 \\
InstructBLIP   & 1582 & 556 \\
MiniGPT-4      & 1169 & 226 \\
Shikra-7B                & 1631 & 446 \\
\bottomrule
\end{tabular}
\caption{
Number of generated object mentions evaluated for the models reported in Table~\ref{tab:hyperparams}. Grounded and hallucinated labels are obtained using the CHAIR object-level evaluation.
}
\label{tab:object_counts}
\end{table}

\begin{table}[t]
\centering
\small
\setlength{\tabcolsep}{3.5pt}
\begin{tabular}{lcccc}
\toprule
Method & Greedy & Beam & Top-\(k\) & Nucleus \\
\midrule
Entropy  & 71.41 & 71.13 & 73.19 & 70.21 \\
NLL      & 70.28 & 59.38 & 70.60 & 67.88 \\
Internal Conf. & 73.88 & 74.90 & 74.38 & 73.30 \\
SVAR     & 74.39 & 73.76 & 71.45 & 73.50 \\
Contextual Lens & 70.93 & 72.97 & 69.59 & 70.55 \\
PAS      & 82.52 & 82.29 & 81.10 & 81.62 \\
GLSim    & 84.09 & 76.53 & 84.47 & 84.31 \\
\midrule
VisER    & \textbf{86.17} & \textbf{82.59} & \textbf{85.72} & \textbf{85.75} \\
\bottomrule
\end{tabular}
\caption{
Effect of decoding strategy on object hallucination detection. %AUROC is reported for LLaVA-1.5-7B on MSCOCO under greedy decoding, beam search, top-\(k\) sampling, and nucleus sampling.
}
\label{tab:decoding_ablation}
\end{table}

\section{Additional Ablation Studies}
\label{app:additional_ablation}

\subsection{Effect of Multi-token Object Representation}

To assess the effect of multi-token aggregation, we compare three strategies: using the first token, the last token, and averaging over all object tokens. Table~\ref{tab:token_aggregation} reports the corresponding results.

\begin{table}[t]
\centering
\small
\setlength{\tabcolsep}{3pt}
\begin{tabular}{lcccc}
\toprule
& \multicolumn{2}{c}{LLaVA-1.5-7B}
& \multicolumn{2}{c}{LLaVA-1.5-13B} \\
\cmidrule(lr){2-3}
\cmidrule(lr){4-5}
Token strategy & AUROC & AUPR & AUROC & AUPR \\
\midrule
First token
& \textbf{86.17} & \textbf{95.37}
& \textbf{83.82} & \textbf{95.35} \\
Last token
& 79.64 & 93.46
& 79.40 & 94.24 \\
Average over tokens
& 85.06 & 95.14
& 83.16 & \textbf{95.35} \\
\bottomrule
\end{tabular}
\caption{Effect of multi-token aggregation on MSCOCO.}
\label{tab:token_aggregation}
\end{table}

As shown in Table~\ref{tab:token_aggregation}, the first-token strategy achieves the highest AUROC on both models, while averaging over all object tokens remains close. These results indicate that VisER is not highly sensitive to the aggregation strategy and support the first-token representation as a simple and effective default.

\subsection{Evidence Gate Distribution}

Since \(M_{\mathrm{vis}}(o)\) is a sum of non-negative visual logit-lens probabilities, the evidence gate satisfies
\[
g(o)\in[0.5,1).
\]
This restricted range is intentional: the gate is designed as a soft evidence validator rather than a standalone detector. The compatibility score already provides useful grounding information, while the gate modulates this compatibility according to object-specific visual evidence rather than completely suppressing it.

Table~\ref{tab:gate_distribution} reports the empirical gate statistics for grounded and hallucinated object mentions.

\begin{table}[t]
\centering
\small
\setlength{\tabcolsep}{3pt}
\begin{tabular}{lccccc}
\toprule
Model
& Grounded
& Halluc.
& Gap \\
& mean
& mean \\
\midrule
LLaVA-1.5-7B
& 0.686 & 0.559 & 0.127  \\
LLaVA-1.5-13B
& 0.667 & 0.549 & 0.118  \\
\bottomrule
\end{tabular}
\caption{Empirical evidence-gate statistics for grounded and hallucinated object mentions on MSCOCO.}
\label{tab:gate_distribution}
\end{table}

As shown in Table~\ref{tab:gate_distribution}, the empirical gate distribution is not degenerate. Grounded objects receive consistently higher mean gate values than hallucinated objects on both model scales. Thus, despite its restricted range, the gate provides meaningful variation in object-specific visual evidence.

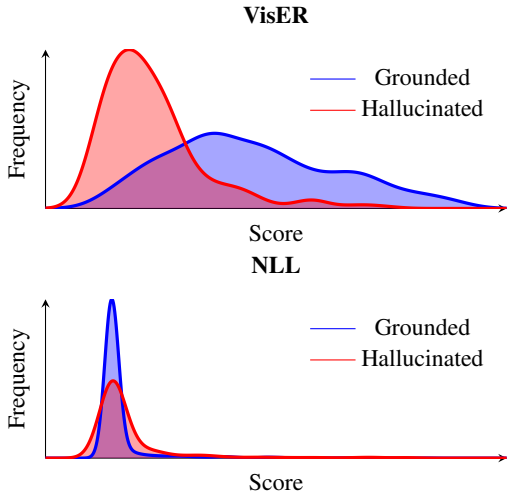
\begin{figure}[H]
\centering
\begin{tikzpicture}
\begin{groupplot}[
    group style={
        group size=1 by 2,
        vertical sep=1.2cm
    },
    width=0.48\textwidth,
    height=0.23\textwidth,
    axis lines=left,
    axis line style={-stealth},
    tick style={draw=none},
    xtick=\empty,
    ytick=\empty,
    ymin=0,
    enlargelimits=false,
    xlabel style={font=\small},
    ylabel style={font=\small},
    title style={font=\small\bfseries},
    legend style={
        draw=none,
        font=\small,
        at={(0.98,0.95)},
        anchor=north east,
        fill=none
    }
]

\nextgroupplot[
    title={VisER},
    xlabel={Score},
    ylabel={Frequency}
]

\addplot[blue, fill=blue, fill opacity=0.35, draw=none]
table[x=x, y=real] {figs/VisER_density.dat} \closedcycle;

\addplot[red, fill=red, fill opacity=0.35, draw=none]
table[x=x, y=hallucinated] {figs/VisER_density.dat} \closedcycle;

\addplot[very thick, blue]
table[x=x, y=real] {figs/VisER_density.dat};

\addplot[very thick, red]
table[x=x, y=hallucinated] {figs/VisER_density.dat};

\legend{Grounded, Hallucinated}

\nextgroupplot[
    title={NLL},
    xlabel={Score},
    ylabel={Frequency}
]

\addplot[blue, fill=blue, fill opacity=0.35, draw=none]
table[x=x, y=real] {figs/nll_density.dat} \closedcycle;

\addplot[red, fill=red, fill opacity=0.35, draw=none]
table[x=x, y=hallucinated] {figs/nll_density.dat} \closedcycle;

\addplot[very thick, blue]
table[x=x, y=real] {figs/nll_density.dat};

\addplot[very thick, red]
table[x=x, y=hallucinated] {figs/nll_density.dat};

\legend{Grounded, Hallucinated}

\end{groupplot}
\end{tikzpicture}
\caption{
Density visualization of object-level scores for grounded and hallucinated objects.
}
\label{fig:score_density}
\end{figure}

\section{Effect of Decoding Strategies}
\label{app:ablation}
%Decoding choices can influence the generation behavior of LVLMs and, consequently, the distribution of hallucinated object mentions. 
Following prior object hallucination detection studies \citep{hoang2026pas, park2025glsim, jiang2025interpreting}, we use greedy decoding as the default setting because it is simple, deterministic, and computationally efficient. To examine whether VisER depends on this specific generation regime, we additionally evaluate detection performance under three alternative decoding strategies, including beam search, top-\(k\) sampling, and nucleus sampling. We use \(N_{\mathrm{beams}}=2\), \(k=10\), and \(p=0.9\) for beam search, top-\(k\), and nucleus decoding, respectively, and apply the same object-level evaluation protocol to the generated captions. As shown in Table~\ref{tab:decoding_ablation}, VisER remains consistently strong across decoding strategies, suggesting that its two-sided grounding score is robust to changes in the generation procedure rather than being specialized to greedy decoding.

\section{Visualization and Examples}

\begin{figure*}[!t]
    \centering
    \includegraphics[width=\textwidth]{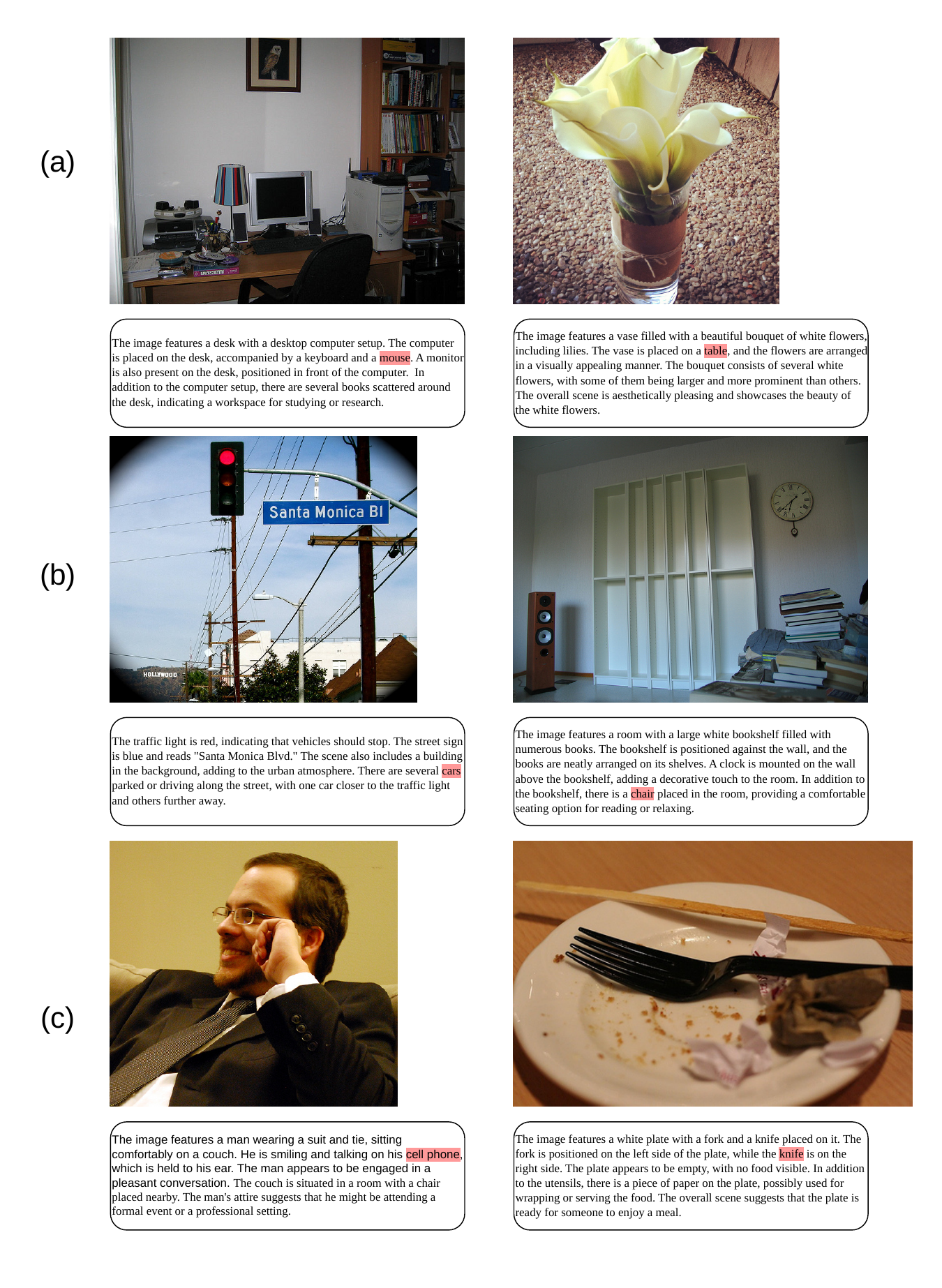}
    \caption{
    Examples of source-confounded hallucinations. 
    Highlighted words indicate hallucinated object mentions in the generated captions. 
    In row (a), both PAS and GLSim incorrectly classify the hallucinated objects as grounded, while VisER correctly detects them as hallucinated. 
    In row (b), GLSim fails but VisER succeeds. 
    In row (c), PAS fails but VisER succeeds. 
    These examples illustrate that baseline support signals can be misled by contextual plausibility or generated-prefix continuation, whereas VisER requires stronger visually sourced evidence.
    }
    \label{fig:qualitative_failures}
\end{figure*}
\usepgfplotslibrary{groupplots}

\begin{figure*}[t]
    \centering
    \includegraphics[width=\textwidth]{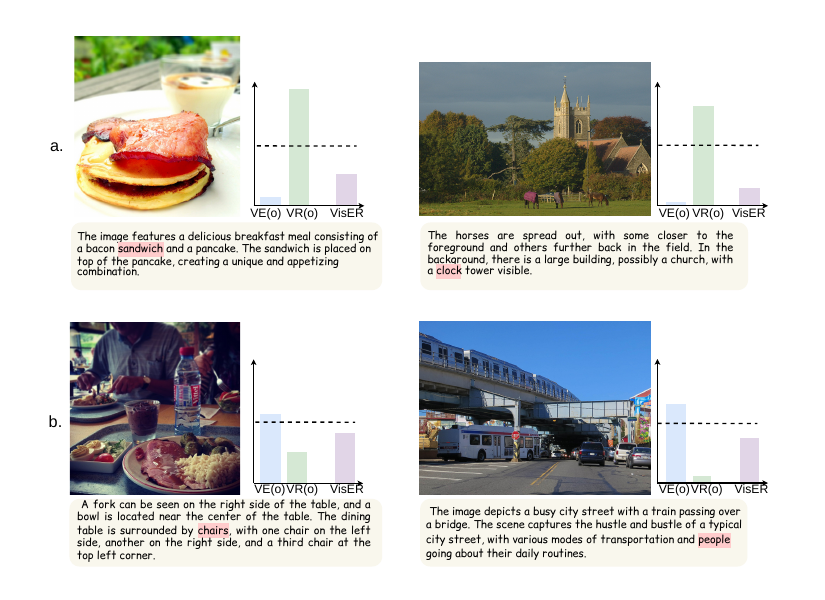}
    \caption{
    Qualitative examples of VisER on hallucinated object mentions. Bars show percentile-normalized Visual Evidence \(VE(o)\), Visual Reliance \(VR(o)\), and VisER; the dashed line denotes the threshold. VisER suppresses hallucinations that are either scene-prior-driven, such as ``sandwich'' and ``clock'', or insufficiently supported by visual evidence, such as ``chairs'' and ``train''.
    }
    \label{fig:qualitative}
\end{figure*}

\subsection{Visualization of Scores}
Figure~\ref{fig:score_density} compares the object-level score distributions produced by NLL and VisER. 
The VisER scores show a substantially clearer distinction between grounded and hallucinated object mentions, 
whereas the NLL distributions remain more concentrated and overlapping. 
This indicates that token likelihood alone provides a limited signal for visual faithfulness, since likely object 
tokens may still be driven by language or scene priors rather than direct image evidence. By contrast, VisER produces a more discriminative scoring space by jointly validating object-specific visual 
evidence and object-level support. 
The resulting separation suggests that VisER reliably distinguishes visually grounded objects from 
mentions induced by contextual or autoregressive priors.

\subsection{Examples of Source-Confounded Support}
\label{app:qualitative}

Figure~\ref{fig:qualitative_failures} provides representative qualitative examples where hallucinated object mentions are correctly detected by VisER but missed by one or more baseline support signals. 
These cases illustrate the source-confounding problem. A hallucinated object can receive high support from scene context, object co-occurrence, or the generated prefix, even when the object is not visually grounded in the image. 
In the first row, both PAS and GLSim, as the best available baselines, incorrectly classify the highlighted hallucinated objects as grounded. 
In the second row, GLSim fails while VisER correctly detects the hallucination.
In the third row, PAS fails while VisER again correctly identifies the highlighted object mentions as hallucinated. 
These examples show that measuring support strength alone can be misleading, whereas VisER improves reliability by checking whether the support is visually sourced.

\subsection{Examples of Scoring}
\label{app:example}
Figure~\ref{fig:qualitative} provides qualitative examples illustrating how VisER combines Visual Evidence \(VE(o)\) and Visual Reliance \(VR(o)\) to detect object-level hallucinations. The bars show percentile-normalized component scores, while the dashed horizontal line denotes the threshold. In the first row, the hallucinated objects are plausible from the scene context: ``sandwich'' is suggested by the breakfast setting, and ``clock'' is suggested by the church-like building. However, both cases receive weak object-specific visual evidence, which lowers the final VisER score despite contextual plausibility. In the second row, the hallucinated objects receive different component patterns: ``chairs'' obtains moderate visual support from the dining scene, while ``train'' is associated with the elevated rail structure, but the final VisER score remains low when the evidence is not sufficiently reliable.

\section{Source-Confounding Analysis}
\label{app:source_confounding}

We provide a simple sufficient-condition analysis showing why the two VisER
components address complementary false-support regimes. Let
\[
M_o=M_{\mathrm{vis}}(o),\qquad
C_o=\mathrm{sim}(h_{\mathrm{ctx}}^{\ell_T},h_o^{\ell_T}),
\]
\[
I_o=S_{\mathrm{img}}(o),\qquad
T_o=S_{\mathrm{text}}(o),
\]
where \(C_o\) denotes object-context compatibility, and let
\(\eta=\tau+\epsilon\). Then
\[
\mathrm{VE}(o)=C_o\sigma(M_o/\eta),
\]
\[
\mathrm{VR}(o)=\frac{I_o}{I_o+T_o+\epsilon}.
\]

\paragraph{Complementary false-support regimes.}
First, consider an evidence-limited hallucination \(o_h\) with nonnegative object-context compatibility \(C_{o_h}\geq0\). The object may be
compatible with the scene or associated with related visual cues, but
object-specific image-token evidence is weak:
\[
M_{o_h}\leq m_h .
\]
Since \(C_{o_h}\geq0\) and \(\sigma(\cdot)\) is monotone,
\[
\mathrm{VE}(o_h)
=
C_{o_h}\sigma(M_{o_h}/\eta)
\leq
C_{o_h}\sigma(m_h/\eta).
\]
Thus, high contextual compatibility alone is not sufficient for a high VE score;
the compatibility term is restricted by object-specific image-token evidence.

Second, consider a prefix-dominated hallucination \(o_h\), where the generated
prefix provides stronger support than the image:
\[
T_{o_h}\geq \kappa I_{o_h},
\qquad \kappa>1 .
\]
For clarity, taking \(\epsilon=0\), we obtain
\[
\mathrm{VR}(o_h)
=
\frac{I_{o_h}}{I_{o_h}+T_{o_h}}
\leq
\frac{1}{1+\kappa}.
\]
Thus, VR is small when support for the object is dominated by the autoregressive
prefix rather than by image-derived evidence.

\paragraph{Source-validation margins.}
Let \(o_g\) be a grounded object and \(o_h\) a hallucinated object. In the
evidence-limited regime, assume
\[
M_{o_g}\geq m_g,\qquad
M_{o_h}\leq m_h,\qquad
m_g>m_h,
\]
with \(C_{o_g} \ge c_g \ge 0\) and \(0 \le C_{o_h} \le c_h\). Then
\[
\mathrm{VE}(o_g)-\mathrm{VE}(o_h)
\geq
c_g\sigma(m_g/\eta)-c_h\sigma(m_h/\eta).
\]
Therefore, VE gives a positive margin whenever
\[
c_g\sigma(m_g/\eta)>c_h\sigma(m_h/\eta).
\]

In the prefix-dominated regime, assume
\[
I_{o_g}\geq \kappa T_{o_g},
\qquad
T_{o_h}\geq \kappa I_{o_h},
\qquad
\kappa>1 .
\]
Again taking \(\epsilon=0\) for clarity,
\[
\mathrm{VR}(o_g)\geq \frac{\kappa}{1+\kappa},
\qquad
\mathrm{VR}(o_h)\leq \frac{1}{1+\kappa},
\]
and hence
\[
\mathrm{VR}(o_g)-\mathrm{VR}(o_h)
\geq
\frac{\kappa-1}{\kappa+1}.
\]

For the final score
\[
s_{\mathrm{VisER}}(o)=
\alpha\mathrm{VE}(o)+(1-\alpha)\mathrm{VR}(o),
\]
define
\[
\Delta_E=\mathrm{VE}(o_g)-\mathrm{VE}(o_h),
\]
\[
\Delta_R=\mathrm{VR}(o_g)-\mathrm{VR}(o_h).
\]
Then we have
\[
s_{\mathrm{VisER}}(o_g)-s_{\mathrm{VisER}}(o_h)
=
\alpha\Delta_E+(1-\alpha)\Delta_R .
\]
Thus, VisER has a positive margin whenever
\[
\alpha\Delta_E+(1-\alpha)\Delta_R>0.
\]
In particular, if \(\Delta_E\geq0\), \(\Delta_R\geq0\), and at least one
inequality is strict, then \(s_{\mathrm{VisER}}(o_g)>s_{\mathrm{VisER}}(o_h)\)
for any \(\alpha\in(0,1)\). This formalizes the complementarity of the two
components: VE is most useful for evidence-limited false support, where
compatibility is not backed by object-specific image evidence, while VR is most
useful for prefix-dominated false support, where the generated prefix explains
the object better than the image.

\section{Usage of Large Language Models}
We used OpenAI GPT-5 language models only for language editing and proofreading. The models were not involved in idea development or method design.
\section{Artifact Licenses}
We use publicly available datasets, models, and baseline methods, including MSCOCO, Pascal VOC, LLaVA, LLaVA-NeXT, InstructBLIP, MiniGPT-4, InternVL3, Shikra, and Qwen2.5-VL, under their respective licenses and terms of use. Our use of these artifacts is limited to academic research and evaluation.

\end{document}